\documentclass[]{spie}  

\usepackage{amsmath,amsfonts,amssymb}
\usepackage{graphicx}
\usepackage[colorlinks=true, allcolors=blue]{hyperref}

\title{Insertion Network for Image Sequence Correspondence Building}

\author[a]{Dingjie Su\textsuperscript{*}}
\author[b]{Weixiang Hong}
\author[a]{Benoit M. Dawant}
\author[a]{Bennett A. Landman}
\affil[a]{Vanderbilt University, Nashville, United States}
\affil[b]{University of Portsmouth, Portsmouth, UK}

\begin{document} 
\maketitle

\begin{abstract}
We propose a novel method for establishing correspondence between two sequences of 2D images. One particular application of this technique is slice-level content navigation, where the goal is to localize specific 2D slices within a 3D volume or determine the anatomical coverage of a 3D scan based on its 2D slices. This serves as an important preprocessing step for various diagnostic tasks, as well as for automatic registration and segmentation pipelines. Our approach builds sequence correspondence by training a network to learn how to insert a slice from one sequence into the appropriate position in another. This is achieved by encoding contextual representations of each slice and modeling the insertion process using a slice-to-slice attention mechanism. We apply this method to localize manually labeled key slices in body CT scans and compare its performance to the current state-of-the-art alternative known as body part regression, which predicts anatomical position scores for individual slices. Unlike body part regression, which treats each slice independently, our method leverages contextual information from the entire sequence. Experimental results show that the insertion network reduces slice localization errors in supervised settings from 8.4 mm to 5.4 mm, demonstrating a substantial improvement in accuracy.
\end{abstract}

\begingroup
\renewcommand{\thefootnote}{\fnsymbol{footnote}}
\footnotetext[1]{dingjie.su@vanderbilt.edu}
\endgroup


\section{INTRODUCTION}
\label{sec:intro}  

This work addresses the problem of building correspondence between two sets of ordered images. Given an image from the first set, the goal is to identify its matching image in the second set, or indicate when no match exists. Solving this problem enables slice-level content navigation in 3D images. For instance, it allows retrieval of 2D slices at key anatomical locations from different subjects, or from different image modalities of the same subject. These matched slices can then be used to build diagnostic models that focus on specific anatomical locations \cite{detection_on_lung_slice, detection_on_liver_slice, detection_on_brain_slice, detection_on_aorta_slice, cnn_rnn_slice_levels, detection_on_cardiac_mri_slice, localization_of_kidney_slice}. Likewise, it can be used to select a subset of volumes that fully capture certain organs, filtering out irrelevant data from large-scale datasets. 

Compared to other image analysis techniques like image registration and semantic segmentation, both of which have seen significant progress in the medical field in recent years \cite{registration_review, segmentation_review}, slice-level content navigation serves as a valuable complement. First, it can answer whether a volume has a complete coverage of a target organ, a question that segmentation or registration cannot reliably answer. Second, many deep learning–based registration and segmentation methods assume particular region-of-interest as input to the network, either in order to improve performance or reduce computation \cite{breast_region_segmentation, wang_atlas-based_2021, ROI_selection}. Slice-level navigation can thus be used to extract such regions. Third, slice-level navigation is significantly faster at the inference stage than many existing segmentation and registration pipelines, making it well-suited for large-scale processing in initial data quality assessment.

In this work, we focus on slice-level content navigation in body CT images along the typical axial direction. A commonly used prior approach for similar tasks is body part regression (BPR) \cite{yan_bpr, tang_bpr, sch_bpr, my_bpr}, where a neural network takes a single slice as input and predicts a position score indicating its anatomical location. This method can be trained in a self-supervised manner by assuming that position scores change linearly with slice indices \cite{yan_bpr, tang_bpr, sch_bpr}, or in a supervised setting using human-annotated ground-truth scores \cite{my_bpr}, which have been shown to improve localization accuracy. However, a key limitation of body part regression is that each slice is processed independently, without considering contextual information from neighboring slices, limiting localization accuracy. This issue has been previously recognized by Tang et al., who introduced a neighboring message passing (NMP) mechanism to incorporate local context \cite{tang_bpr}. However, their approach uses a limited window size, and both their results and ours show that the performance gain from NMP is limited.

Here, we present a new approach to slice-level navigation by building slice-to-slice correspondence between a template and target slice sequences. Inspired by attention-based models in natural language processing, we represent a 3D volume as a sequence of 2D slices and use attention mechanisms to establish correspondence between volumes by inserting slices from one volume into another. This formulation introduces two main challenges. First, the network must effectively capture both the visual features within each slice and the contextual dependencies across slices. Second, modeling the insertion behavior and designing an appropriate loss function is nontrivial. An L1 loss between the true and predicted insertion positions is unsuitable due to the non-differentiability of the argmax operation. At the same time, a standard cross-entropy loss based on the one-hot encodings of ground-truth positions ignores the ordinal structure inherent to slice positions. To address the first challenge, we adopt a hybrid architecture that combines a CNN \cite{cnn} for slice feature extraction with a transformer \cite{attention_is_all_you_need} for modeling cross-slice dependencies. To address the second challenge, we represent the ground-truth insertion position not as an integer but as a Gaussian distribution centered at the integer position. We then use a distance metric between probability distributions as the loss function. We compare to body part regression techniques and demonstrate improved performance.

\section{METHOD}

   \begin{figure} [ht]
   \begin{center}
   \begin{tabular}{c} 
   \includegraphics[width=\textwidth]{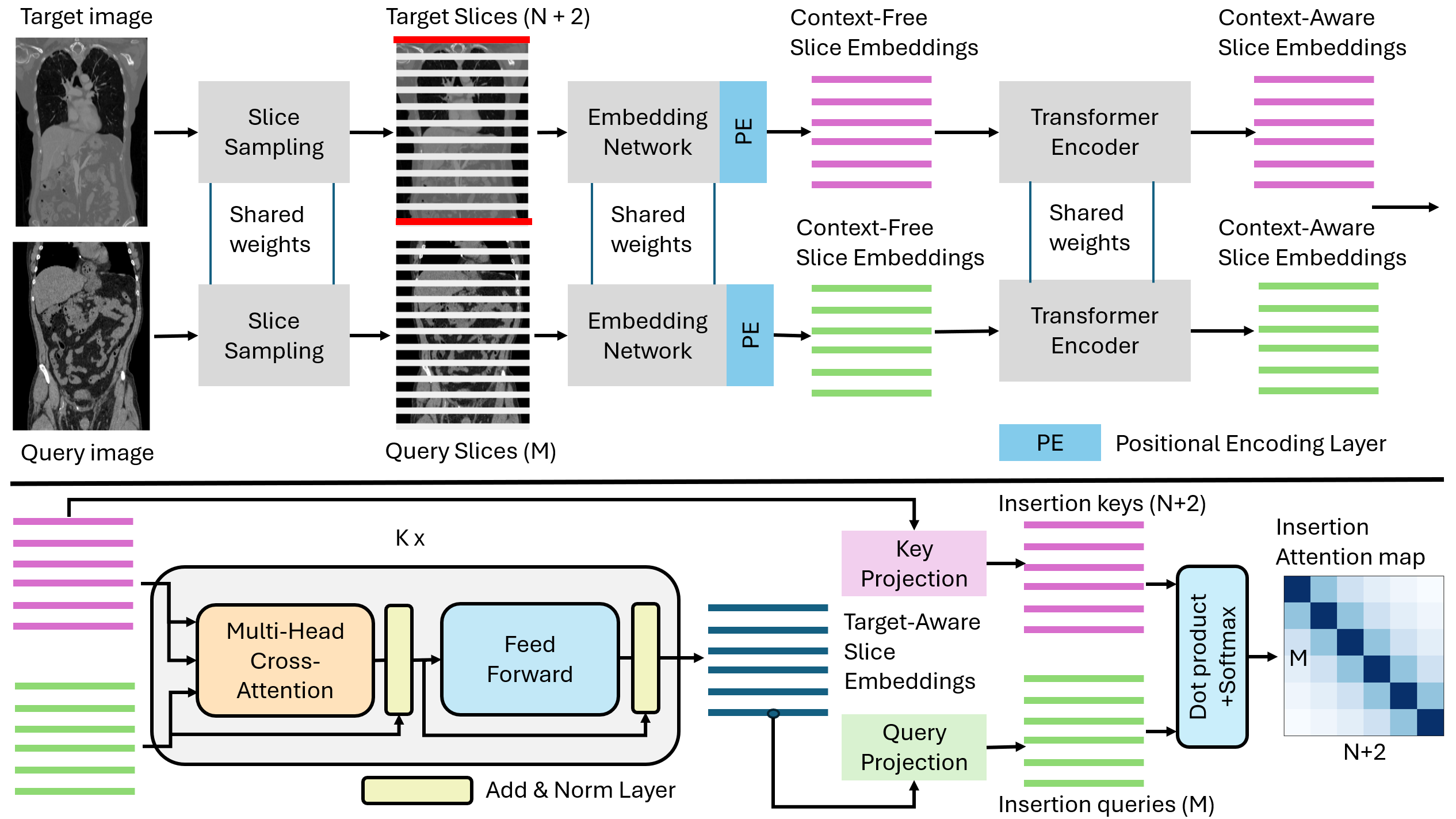}
   \end{tabular}
   \end{center}
\caption[fig1] 
   { \label{fig1} 
Insertion network for inserting slices from the query image into the target image. The top panel illustrates the encoding phase, while the bottom panel shows the insertion phase. Both phases are connected and trained jointly in an end-to-end manner. The target image contains N+2 slices, where two artificial slices are added at the beginning and end to represent insertion positions outside the original slice range. The embedding network uses a ResNet18 which produces 512-dimensional slice emebddings.}
   \end{figure}

\subsection{Method overview}

Given two sequences of elements, $A = (a_1, a_2, \ldots, a_M)$ and $B = (b_1, b_2, \ldots, b_N)$, our goal is to identify which element in $B$ is similar to an element in $A$. Note the importance of the context here because comparing elements in isolation is often ill-defined. For example, we cannot say that 3 is similar to 5, but we can say that 3 in the sequence $(3, 9, 15, 21, 27)$ is similar to 5 in the sequence $(5, 13, 20, 27, 34)$ because both are the smallest elements in their respective sequences. The same logic applies to medical medical images. For instance, when identifying slice at certain anatomical locations (e.g., lower end of kidney), human labelers typically consult the coronal or sagittal views—which provide anatomical context—rather than the axial view, where slices appear in isolation.

In this work, we establish such contextual correspondence by modeling the task as an insertion problem: we determine where each element in $A$ should be inserted within $B$. This formulation allows us to relax the requirement of a direct element-to-element matching between $A$ and $B$. To implement this, we convert $A$ into a sequence of queries $Q = (q_1, \ldots, q_M)$ and $B$ into a sequence of keys $K = (k_1, \ldots, k_N)$, and compute similarity between each query and key using the dot product. Figure \ref{fig1} illustrates the architecture we propose for transforming raw image slices into queries and keys. First, an embedding network extracts features from individual 2D slices without considering their context. CNNs are preferred at this stage due to their strong inductive bias toward local features. These context-free embeddings are then passed through a transformer encoder, which uses self-attention to produce context-aware slice embeddings. Subsequently, $K$ layers of cross-attention are applied between the sequences, where queries are derived from the query image and keys/values from the target image. Finally, learnable projection matrices are applied to transform the target slice embeddings and the context-enriched query embeddings into the final sets of keys and queries, respectively. In our experiments, we use ResNet-18 \cite{resnet} as the Embedding Network, which produces 512-dimensional embedding vectors. The Transformer Encoder consists of two self-attention layers with 8 attention heads each, followed by two cross-attention layers with 8 heads for subsequent processing.

The following sections describe the technical details of the proposed method.

\subsection{Slice sampling}
In practice, the query and target sequences may contain different numbers of elements, corresponding to CT volumes with varying numbers of slices. To ensure the input fits in GPU memory, we sample up to 256 slices for the Embedding Network shown in Fig. \ref{fig1}. For volumes with fewer than 256 slices, we pad with zeros; for volumes with more than 256 slices, we perform sampling. We use two sampling strategies: the first uniformly samples slices across the entire volume at regular intervals, ensuring coverage of the whole volume. The second randomly selects a subvolume and samples slices from it. It serves as a data augmentation technique that generates multiple partial views from the same volume. During inference, we always use uniform sampling as we want to locate slices within the full volume. During training, we randomly apply either strategy to expose the model to more diverse data.

\subsection{Ground-truth insertion position and loss function}

We adopt supervised training by providing ground-truth insertion positions. They are obtained under the assumption that the human body exhibits a piecewise linear anatomical arrangement. Specifically, we manually label seven key axial slices in each CT volume, as shown in Fig. \ref{ks}. Each key slice is assigned a fixed position score, and the position scores for slices between key slices are obtained via linear interpolation. The insertion position for a query slice is then determined by the location in the target image with the closest position score. To handle cases where the query and target images have different fields of view, we introduce an artificial starting slice (filled with ones) before the most superior slice of the target volume and an artificial ending slice (filled with negative ones) after the most inferior slice. Query slices that are more superior than the target volume are inserted at the position of the artificial starting slice, while query slices that are more inferior are inserted at the position of the artificial ending slice.

After obtaining the ground-truth insertion position, we represent it as a Gaussian distribution centered at that position. The variance of the Gaussian controls the rate at which the probability decays and reflects our confidence in the insertion position: higher confidence leads to faster decay, thereby penalizing even small deviations more strongly. In this work, we use a fixed variance of 25 for all key slices, corresponding to a standard deviation of 5 mm, which approximates the expected human annotation error for this key-slice localization task. For slices near the boundaries, the distribution is truncated and renormalized to sum to 1. 

To enable comparison with the ground-truth distributions, we normalize the predicted attention weights to obtain a probability distribution. In this work, we consider two commonly used probability distance metric as the loss function: the Kullback–Leibler (KL) divergence and the Earth Mover’s Distance (EMD). Empirically, we find that KL divergence performs better for this task (see Fig. \ref{fig4}).

The KL divergence is defined in Equation \ref{eq:kl}, where $P_i$ and $Q_i$ denote the ground-truth and predicted insertion probabilities at position $i$, respectively. When used as a loss function, Equation \ref{eq:kl} can be simplified to Equation \ref{eq:kl_loss} because the ground-truth probability $P$ is a fixed Gaussian. As a result, its entropy term $\sum_{i=1}^{N} P_i \log(P_i)$ is constant and can be omitted from the loss. The EMD is defined in Equation \ref{eq:emd} and is used directly as the loss without further simplification. 

\begin{equation}
\label{eq:kl}
D_{\mathrm{KL}}(P \,\|\, Q) = \sum_{i=1}^{N} P_i \, \log \frac{P_i}{Q_i}
\end{equation}

\begin{equation}
\label{eq:kl_loss}
{\mathrm{Loss_{KL}}}(P, Q) = -\sum_{i=1}^{N} P_i \, \log Q_i
\end{equation}

\begin{equation}
\label{eq:emd}
\mathrm{Loss_{EMD}}(P, Q)
=
\sum_{k=1}^{N}
\left|
\sum_{i=k}^{k} (P_i - Q_i)
\right|
\end{equation}
 
\subsection{Positional encoding for slice embeddings}
We add positional encodings to the slice embeddings produced by the Embedding Network before feeding them into the Transformer Encoder. The positional encoding follows the standard sinusoidal formulation, as defined in Equation \ref{eq:pe}.

Based on Equation \ref{eq:pe}, there are two choices for the position variable $pos$. The first uses the natural slice index, $pos=1,2,3,..,K$. This encoding is independent of the slice spacing and is therefore referred to as absolute positional encoding. The second scales the natural index by the slice spacing (in millimeters) of the corresponding CT volume. Because this encoding depends on the slice spacing, we refer to it as relative positional encoding. We evaluated both approaches empirically and found that absolute positional encoding yields better performance for this task (see Fig. \ref{fig4}).

\begin{equation}
\label{eq:pe}
\mathrm{PE}(pos, k) =
\begin{cases}
\sin\left(\dfrac{pos}{10000^{\frac{k}{d}}}\right), & k \text{ even} \\
\cos\left(\dfrac{pos}{10000^{\frac{k-1}{d}}}\right), & k \text{ odd}
\end{cases}
\end{equation}

where $pos$ is the position index, $k$ is the dimension index, d is the embedding dimension.

\subsection{Dataset}

\begin{figure} [ht]
   \begin{center}
   \begin{tabular}{c} 
   \includegraphics[width=\textwidth]{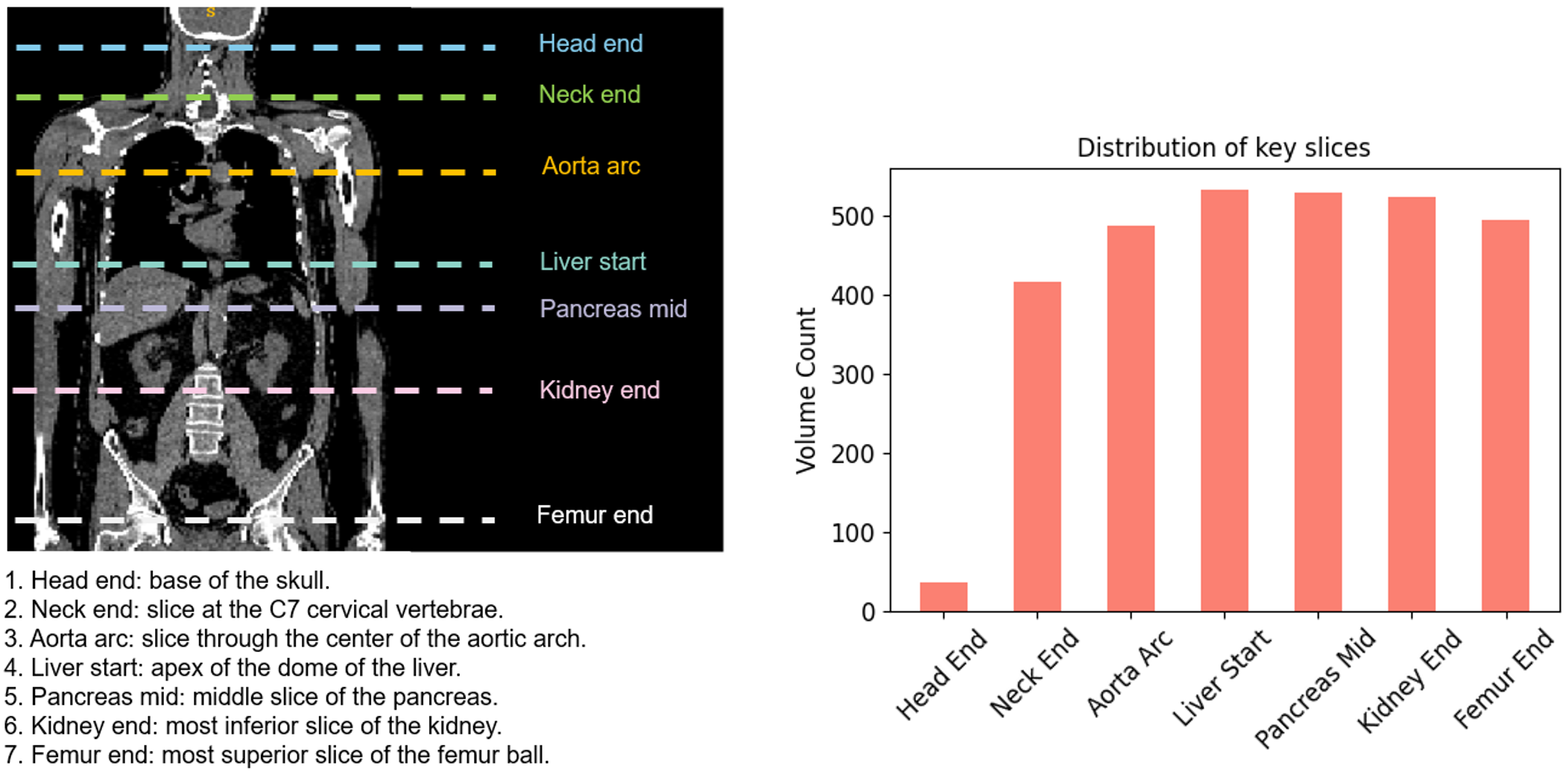}
   \end{tabular}
   \end{center}
\caption[ks]
   { \label{ks} Left: Coronal view of a body CT with 7 manually labeled key slices. Right: Total number of key slices labeled in our dataset.}
   \end{figure}

Our dataset consists of 533 body CT volumes acquired at Vanderbilt University Medical Center using diverse scanner types. The total number of key slices labeled in these volumes is shown in Fig. \ref{ks}. Among the 533 volumes, 400 volumes are used for training, 33 for validation, and 100 for testing. The volume dimensions range from 512 × 512 × 167 to 512 × 512 × 1379 voxels, where the last dimension indicates the number of slices. The voxel resolution ranges from 0.58 × 0.58 × 0.40 mm to 1.50 × 1.50 × 3.00 mm. Before being fed into the network, all slices are downsampled to a spatial resolution of 256 × 256, while the slice spacings are kept at their original values. All data were retrieved in de-identified form under IRB approval for human data at Vanderbilt University (IRB number: 140274).

\section{RESULTS}

We train the insertion network for 100 epochs using the Adam optimizer with a learning rate of 1e-4. The best epoch is chosen according to validation accuracy. 

\begin{figure} [ht]
   \begin{center}
   \begin{tabular}{c} 
   \includegraphics[width=\textwidth]{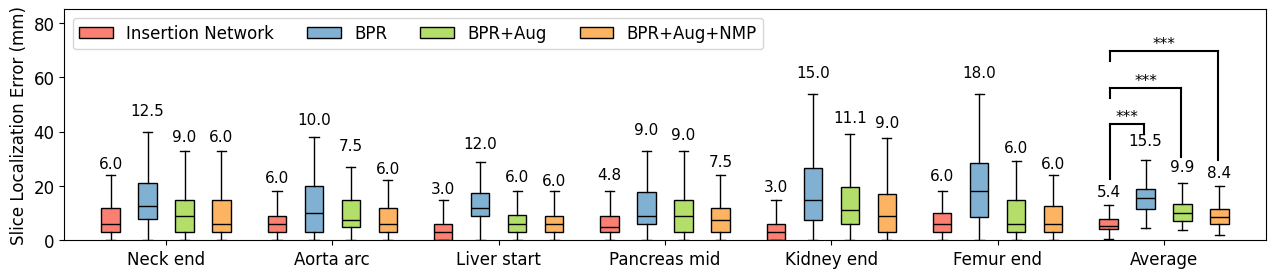}
   \end{tabular}
   \end{center}
\caption[fig3]
   { \label{fig3} Quantitative comparison of the slice location errors. Wilcoxon test was used for statistical analysis $(***: p < 0.001)$. The Head end location is excluded from the plot because the number of cases at this location is insufficient for reliable statistical analysis.}
   \end{figure}

In Figure \ref{fig3}, we present a quantitative analysis of key slice localization errors and compare our method to body part regression techniques. The BPR models are trained in a supervised manner, following the approach in [\cite{my_bpr}], using the same ground-truth position scores that were used to derive the ground-truth insertion positions. We evaluate three BPR variants: (1) BPR, a baseline network trained on raw images without advanced data augmentation; (2) BPR+Aug, based on the method in [\cite{my_bpr}], which introduces a selective mix-up augmentation to promote position score invariance across subjects at the same anatomical location; and (3) BPR+Aug+NMP, which combines selective mix-up with the Neighbor Message Passing (NMP) technique from [\cite{tang_bpr}] to incorporate limited slice-level contextual information. As shown in the `Average' column of Figure \ref{fig3}, which is the average error across all key slices, both selective mix-up and NMP reduce localization error compared to the baseline. However, even the best-performing BPR variant performs significantly worse than our insertion network. Notably, the insertion network operates directly on raw images without requiring any augmentation. Its superior performance stems from the richer context it leverages: when inserting a slice, the network considers not only the full content of the query image but also the entire target image, enabling more informed and accurate decisions than BPR-based approaches.

\begin{figure} [ht]
   \begin{center}
   \begin{tabular}{c} 
   \includegraphics[width=\textwidth]{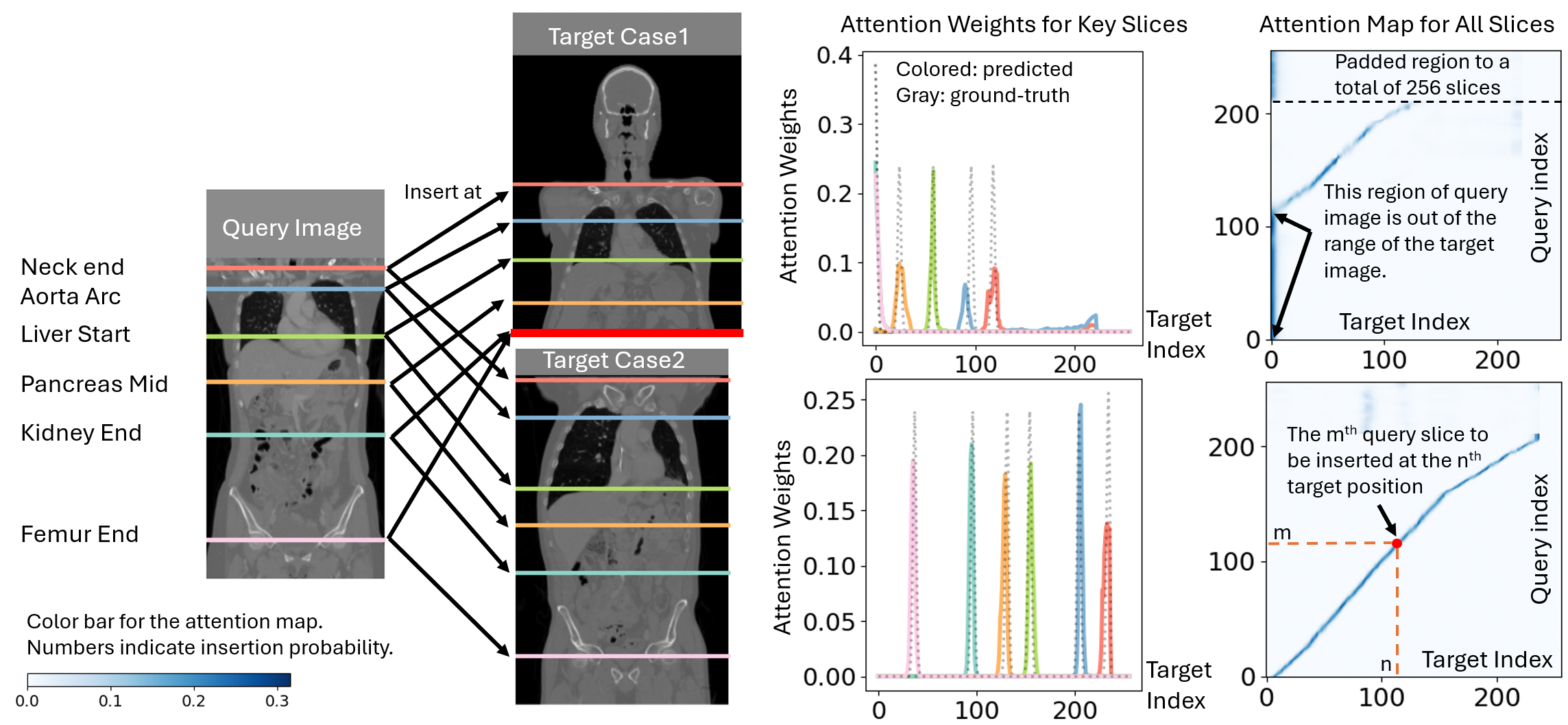}
   \end{tabular}
   \end{center}
\caption[fig2] 
   { \label{fig2} Visual demonstration of the insertion behavior on 2 test cases having different FOVs. The red line at the bottom of target case 1 represents the artificial ending slice that query slices can point to in order to indicate they are “out of range”. }
   \end{figure}

Figure \ref{fig2} presents two visual examples illustrating the insertion behavior of the proposed method. In both test cases, the network correctly inserts key slices that are shared between the query and target images, despite noticeable differences in anatomical shape and relative spatial location. In test case 1, the target image has a substantially different field of view from the query image. The network successfully handles this by inserting all query slices below the kidney—which are absent in the target image—into the position of the artificial ending slice (shown as the red line at the bottom of the target image) used to indicate out-of-range locations. The attention maps shown in the final column demonstrate that the query slices are inserted into the target image in an approximately linear order. This behavior is consistent with the linear assumptions underlying body part regression methods and with the assumptions used when defining the ground-truth insertion positions.

\begin{figure} [ht]
   \begin{center}
   \begin{tabular}{c} 
   \includegraphics[width=\textwidth]{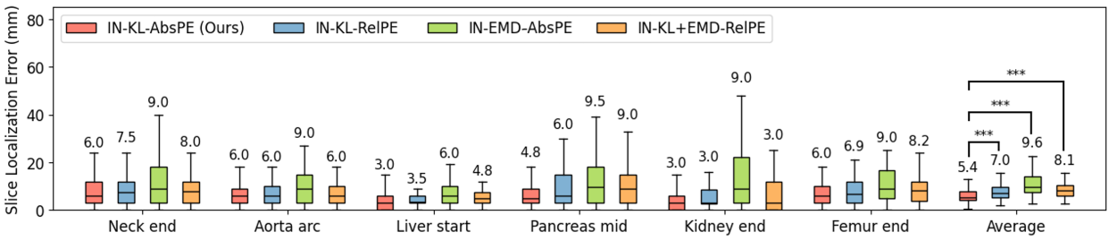}
   \end{tabular}
   \end{center}
\caption[fig4]
   { \label{fig4} Insertion network trained with different positional encodings and loss functions. IN: insertion network. KL: KL divergence. EMD: earth mover's distance. AbsPE: absolute positional encoding. RelPE: relative positional encoding. The Head end location is excluded from the plot because the number of cases at this location is insufficient for reliable statistical analysis.}
   \end{figure}

In Figure \ref{fig4}, we present ablation studies evaluating the performance of KL divergence and EMD, and the choice of absolute and relative positional encodings. The results show that using absolute positional encoding (based on natural index only) together with KL divergence as the loss function yields the best results.

\section{Conclusion}
We propose a novel method for establishing correspondences between two image sequences and demonstrate its effectiveness on a slice navigation task. To this end, we introduce a sequence-to-sequence comparison model that captures the full context of both the query and target sequences. This context-awareness substantially improves slice navigation accuracy, outperforming the previous method, body part regression, which addresses the same problem. Unlike traditional sequence modeling methods, our approach uses a novel learning strategy that applies supervision directly to the attention weights rather than the network outputs.

In clinical scenarios, an important application of slice navigation is image preprocessing for excluding images that contain unwanted anatomical regions. In this context, inference speed is a key advantage of slice-level navigation compared with other approaches such as full volume segmentation. When run on an RTX 3090 GPU, the insertion network processes a volume with 300 slices in approximately 0.98 seconds, requiring about 2.7 hours to process 10,000 such volumes. In contrast, a popular segmentation method, TotalSegmentator \cite{totalseg}, requires approximately 360 hours to process the same number of volumes, making it an unnecessary—and possibly prohibitive—computational investment for image preprocessing alone.

While the current work focuses on spatial sequences, the same network architecture can be extended to temporal sequences—for example, to analyze disease progression from longitudinal data, which is an interesting future direction.

\section*{ACKNOWLEDGEMENT}
This work is funded by the National Center for Advancing Translational sciences (NCATS) Clinical Translational Science Award (CTSA) Program under award number 5UL1TR002243-03. The authors gratefully acknowledge National Institutes of Health HuBMAP grants U54DK134302 and U54EY032442. The content is solely the responsibility of the authors and does not necessarily represent the official views of the National Institutes of Health. This work was conducted in part using the resources of the Advanced Computing Center for Research and Education at Vanderbilt University.

\bibliography{report} 
\bibliographystyle{spiebib} 

\end{document}